%% file: article_main.tex
\documentclass[fleqn,10pt]{wlscirep}

\usepackage{authblk}
\usepackage{capt-of}
\usepackage{subfigure}
\usepackage{wrapfig}
\usepackage{hyperref}       
\usepackage{url}            
\usepackage{booktabs}       
\usepackage{amsfonts}       
\usepackage{nicefrac}       
\usepackage{microtype}      
\usepackage{lipsum}
\usepackage{listings}
\usepackage{fancyhdr}       
\usepackage{graphicx}       
\usepackage{chemformula}
\graphicspath{{png/}}    
\usepackage[section]{placeins}
\usepackage{tabularx}
\usepackage{subfigure}
\usepackage[utf8]{inputenc}
\usepackage[T1]{fontenc}
\usepackage{caption}
\usepackage{booktabs}
\usepackage{float}
\usepackage{parskip}
\usepackage{hyperref}
\usepackage{multirow}
\usepackage{tablefootnote}
\usepackage{threeparttable}
\usepackage{chemformula}
\usepackage{xcolor}
\usepackage[section]{placeins}
\usepackage{setspace}
\begin{document}
\title{The Battle of Information Representations: Comparing Sentiment and Semantic Features for Forecasting Market Trends}

\author[1]{Andrei ZAICHENKO}
\author[2]{Aleksei KAZAKOV}
\author[2]{Elizaveta KOVTUN}
\author[2,3,*]{Semen BUDENNYY}

\affil[1]{Higher School of Economics University, Moscow}
\affil[2]{Sber AI Lab, Moscow}
\affil[3]{Artificial Intelligence Research Institute (AIRI), Moscow}

\affil[*]{Corresponding author: budennyysemen@gmail.com }

\begin{abstract}

The study of the stock market with the attraction of machine learning approaches is a major direction for revealing hidden market regularities. This knowledge contributes to a profound understanding of financial market dynamics and getting behavioural insights, which could hardly be discovered with traditional analytical methods. Stock prices are inherently interrelated with world events and social perception. Thus, in constructing the model for stock price prediction, the critical stage is to incorporate such information on the outside world, reflected through news and social media posts. To accommodate this, researchers leverage the implicit or explicit knowledge representations: (1) sentiments extracted from the texts or (2) raw text embeddings. However, there is too little research attention to the direct comparison of these approaches in terms of the influence on the predictive power of financial models. In this paper, we aim to close this gap and figure out whether the semantic features in the form of contextual embeddings are more valuable than sentiment attributes for forecasting market trends. We consider the corpus of Twitter posts related to the largest companies by capitalization from NASDAQ and their close prices. To start, we demonstrate the connection of tweet sentiments with the volatility of companies' stock prices. Convinced of the existing relationship, we train Temporal Fusion Transformer models for price prediction supplemented with either tweet sentiments or tweet embeddings. Our results show that in the substantially prevailing number of cases, the use of sentiment features leads to higher metrics. Noteworthy, the conclusions are justifiable within the considered scenario involving Twitter posts and stocks of the biggest tech companies. 




\end{abstract}

\maketitle

\begin{keywordname}
stock market, embedding, time series, event study, NLP
\end{keywordname}
\input{sections/intro_kazakov}

\input{sections/contribution}
\input{sections/related_works_kazakov}


\input{sections/methods}

\input{sections/results}

\input{sections/conclusions_revised}
\newpage
\bibliography{source.bib}

\newpage
\input{sections/annex}

\end{document}

%% file: sections/intro_kazakov.tex
\section{Introduction}

Stock price prediction is a challenging problem in the financial domain that draws significant interest from researchers. There is a massive amount of groundwork on machine learning applications for forecasting financial market trends. The developed methods could be divided into two streams: fundamental and technical analysis. 

Technical analysis relies solely on historical structured data of stock markets. Such analysis is extensively used with machine learning techniques
\cite{AYALA2021107119, PENG2021100060}. These studies demonstrate that technical indicators such as Exponential Moving Average (EMA) and Moving Average
Convergence/Divergence (MACD) can be used to increase the profitability of trading signals.

In contrast, fundamental analysis focuses on any useful information outside of historical market data regarding
the stock in question, such as the financial environment, law regulations, social networks, geopolitical stability, and news. 
While both approaches are usually used separately, recent studies show that a combination of the two can yield more accurate predictions \cite{Nti2020}.

With the advancements in machine learning and natural language processing (NLP), researchers explore the leveraging
of various features extracted from textual data for predicting stock prices. Such NLP technique as sentiment analysis allows the extraction of the sentiment or emotion from a piece of text, which can be used to infer the overall sentiment of the market towards a particular company or stock. 
Information representation in the context of financial markets is proposed to be viewed as explicit or implicit. 

The implicit representation involves extracting sentiment polarity directly from text (by a pre-trained NLP classifier) and then using this information to assess the expected reaction of a signal. This typically involves classifying the text into positive or negative categories based on its perceived sentiment towards market and using this classification as a feature in a further machine learning model to predict price movements. This approach has the advantage of being relatively simple and straightforward to implement, but it has the drawback of losing important semantic features and contextual information in the text. Moreover, the implicit approach relies on the product of the used sentiment analysis algorithm, whose inaccurate operation can introduce distortions into the work of the final predictive model. 

The explicit approach involves creating an embedding from the text directly and then using this embedding to predict the reaction for signal change \cite{10.1007/978-3-030-34223-4_5}. An embedding is a mathematical representation of a text in a lower-dimensional space, which can capture the meaning and context of the word or phrase. By creating an embedding from text, this approach is able to retain more of the semantic links contained in the text, which may be useful for predicting financial market trends. However, this approach may be more complex and time-consuming to implement, as it involves creating and training an embedding model on the text data.

After analyzing two ways of representing information related to the stock market, we can conclude that each of them has its own strengths and weaknesses. In this work, our aim is to compare the performance of the stock price prediction model built on the basis of either implicit or explicit knowledge representations.

%% file: sections/contribution.tex
\section{Contribution}

In this research, our goal is to explore the effectiveness of the explicit embedding vector approach and compare it with the more established implicit sentiment solution. The main contributions achieved during the study of this topic are as follows:

\begin{itemize}
    \item Demonstrated the statistical dependence between stock price volatility and Twitter post sentiments.
    \item Proved intrinsic linkage between two kinds of representations 
    by showing that embeddings hold information on sentiments.
    \item Discovered the superiority of the sentiment extraction approach over the usage of embedding vectors in the prevailing number of cases. 
\end{itemize}

%% file: sections/related_works_kazakov.tex
\section{Related work}

One of the main streams of the research is dedicated to studying the usage of purely technical analysis trading indicators and historical data in combination with statistical methods for stock price prediction with machine learning. Many researchers employ the GRU-based models in this tasks \cite{Aseeri_2023, app13010222, GUPTA2022117986} while the others explore the transformer architecture in this field \cite{WANG2022118128, numhtml, 10027785}. The goal of such research is to provide investors and other financial actors with an insight of price movements to make more informed financial decisions relying on technical analysis as the foundation.

With the development of NLP methods, the number of works aiming to predict stock price trends and volatility proposed a combination of financial news and social media data is increasing \cite{Khan2022}. \cite{10.1145/3159652.3159690} pointed out the lack of trustworthiness and comprehensiveness of online content collected from social media and low quality news sources. Clinical trial announcements were used as a source of sentiment in pursuit of predicting pharma stock market price changes by \cite{https://doi.org/10.48550/arxiv.2208.07248}.  Another research used Valence Aware Dictionary and Sentiment Reasoning (VADER) \cite{hutto2014vader} for sentiment analysis \cite{math10122001}. \cite{Li2022} proposed a novel Deep Learning Transformer Encoder Attention (TEA) model. \cite{https://doi.org/10.48550/arxiv.2005.02527} considered such under-explored content as Environmental, Social, and Corporate Governance (ESG) news flow for volatility forecasting. \cite{AUDRINO2020334} analyzed the impact of sentiment and attention variables on the stock market volatility by adding search engine and information consumption data on top of widely used social media and news texts.

Sentiment analysis involves extracting the sentiment
or emotion from a piece of text, which can be used to infer the overall sentiment of the market towards a particular company
or stock. Sentiment scores are commonly used in stock price prediction studies \cite{Khan2022, 10.1145/3159652.3159690, math10122001, Li2022, https://doi.org/10.48550/arxiv.2005.02527, AUDRINO2020334} as they are easy to compute and provide a
simple metric to gauge market sentiment.
On the other hand, embedding vectors are dense numerical representations of words or phrases that capture the semantic
meaning of the text. They are created by mapping words or phrases to high-dimensional vectors in a way that similar words or phrases are located close to each other in this vector space using techniques like Word2vec \cite{https://doi.org/10.48550/arxiv.1301.3781} or GloVe \cite{pennington-etal-2014-glove}, BERT \cite{https://doi.org/10.48550/arxiv.1810.04805} and GPT \cite{https://doi.org/10.48550/arxiv.2202.08904}, and have been shown to
be effective in capturing complex relationships between words and phrases. 

There are several works that concentrate on the usage of text semantics in the context of stock price prediction. \cite{10.1007/978-3-030-34223-4_5} proposed a Multi-head Attention Fusion Network to exploit aspect-level semantic information from texts to enhance prediction. \cite{Lin2022} developed a Spatial-temporal attention-based convolutional network. The authors converted news articles into 300-dimensional vector embeddings and used them as a feature in their model. They noted that in the case of the utilization of the preprocessed text features, latent information in the text is lost because the relationships between the text and stock price are not considered. \cite{Chandola2022} employed Word2Vec and LSTM algorithms, while \cite{Chen2022} adopted a transformer architecture using high level textual features. The main conclusion derived from these papers can be formulated as follows - the effectiveness of exploiting and fusing semantic aspect-level textual information leads to an improved performance upon the baselines. This fact means that the topic needs further investigation and refining.


The main advantage of embeddings over sentiment analysis is that they capture more complex relationships in the data. Despite the promising results of using embedding vectors as a feature \cite{Chandola2022, Chen2022}, there is still a research gap in the comparison of the effectiveness of using embedding vectors versus sentiment scores in stock price prediction. Moreover, reserachers tend to experiment with different datasets and methodologies, making it difficult to draw meaningful comparisons.



%% file: sections/methods.tex
\section{Methods}

\begin{figure}[!ht]
  \includegraphics[scale=0.5]{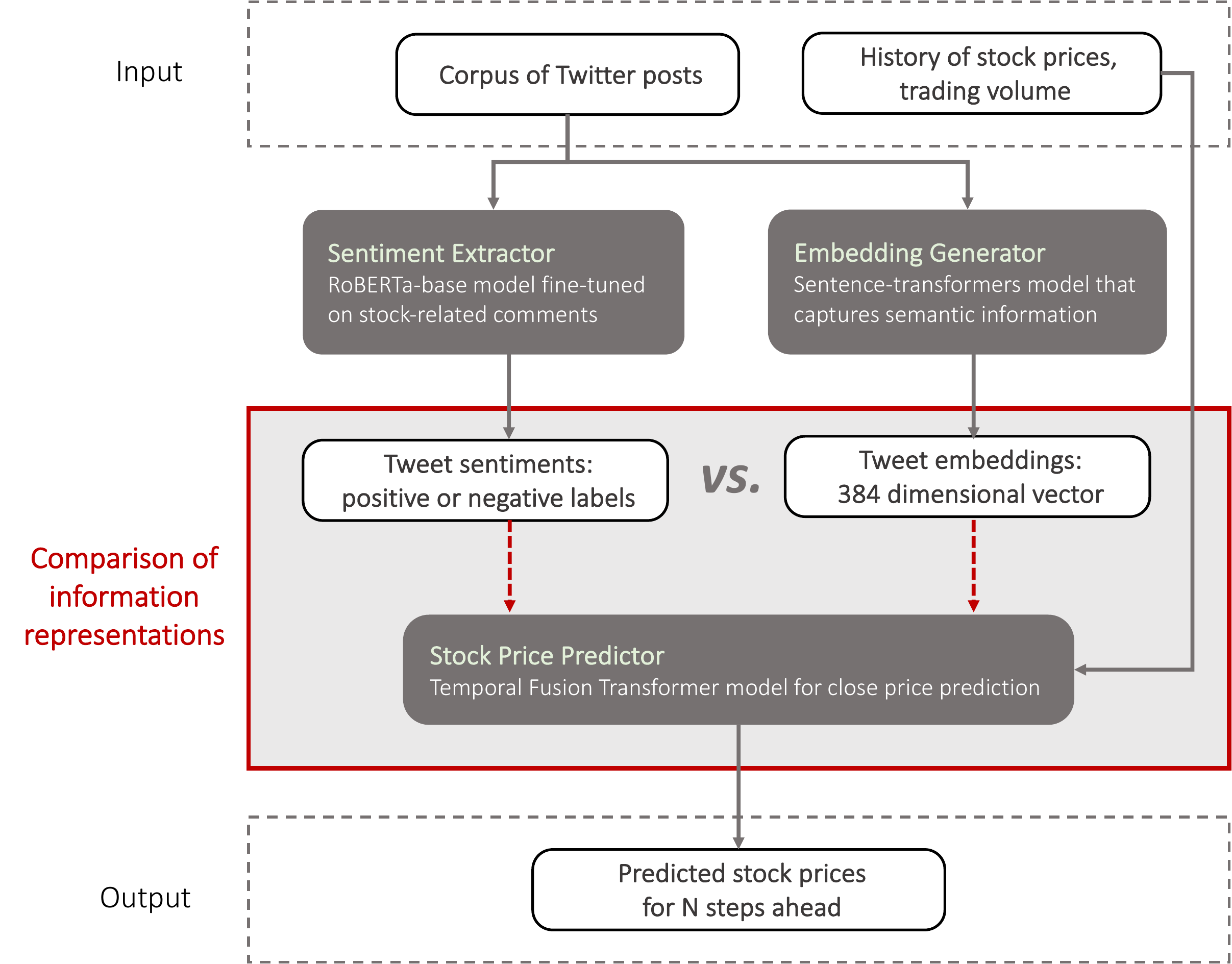}
  \centering
  \caption{Pipeline for comparison of utilizing either text sentiments or embeddings within the stock price prediction problem statement. There are three main steps in the scheme: (1) preprocessing of the collected Twitter and financial datasets, (2) obtaining either tweet sentiments or tweet embeddings, and (3) making prediction of stock close prices for $N$ steps ahead.}
  \label{fig:schema}
\end{figure}

In this section, we highlight the key steps of our pipeline for constructing the model for stock price prediction. In particular, we focus on the phase in which we use different techniques for getting textual information representations that served as part of the model input. We emphasize the prediction quality depends on the formed feature space and provide a framework for fair comparison and selection of the optimal configuration. An overall scheme of processes to make a stock close price prediction is described in Figure \ref{fig:schema}. It consists of three main steps:

\begin{itemize}
    \item \textbf{Data retrieval and preprocessing}. As the initial data, we take a corpus of Twitter posts related to the selected group of companies and the historical values of their stock prices and trading volumes. Tweets are subject to data cleaning. Financial data are modified through the volume feature being smoothed. After all, a business day resampling is applied to get our attributes at the same time frequency. 
    \item \textbf{Model input formation}. Depending on the experiment, we pass the Twitter dataset either through the sentiment extractor or embedding generator to get the part of the input, which is responsible for the outside world information. Calendar holidays, trading volumes, and preceding close prices constitute the rest part.
    \item \textbf{Model prediction}. Data samples described with the composed feature space are fed into the TFT model. After a training process, a $N$ steps ahead prediction for the close price is made. 
\end{itemize}

\subsection{Data preparation}
To get accurate price prediction for the next $N$ steps, we need to properly prepare initial data. We construct the predictions on the basis of Twitter and financial datasets. From tweets, we can get either sentiment or semantic information. Also, we consider historical financial information as a part of model input. Moreover, for further deep market analysis, we provide the formula for the price volatility evaluation. 

\subsubsection{Data}\label{data}

For Twitter data, we use the dataset from Kaggle published in 2020 \cite{9378170}. This dataset contains tweets related to 5 companies: Amazon, Apple, Google, Microsoft, and Tesla from 2015 to 2020. Raw data contains over 3 million tweets and information on the tweet author, post date, tweet text body, and the number of comments, likes, and retweets. For the preprocessing steps, a simple spam and duplicate tweet reduction was performed. All the preprocessing steps in more detail are described in the Appendix.

\begin{figure}[!ht]
  \includegraphics[width=0.7\textwidth]{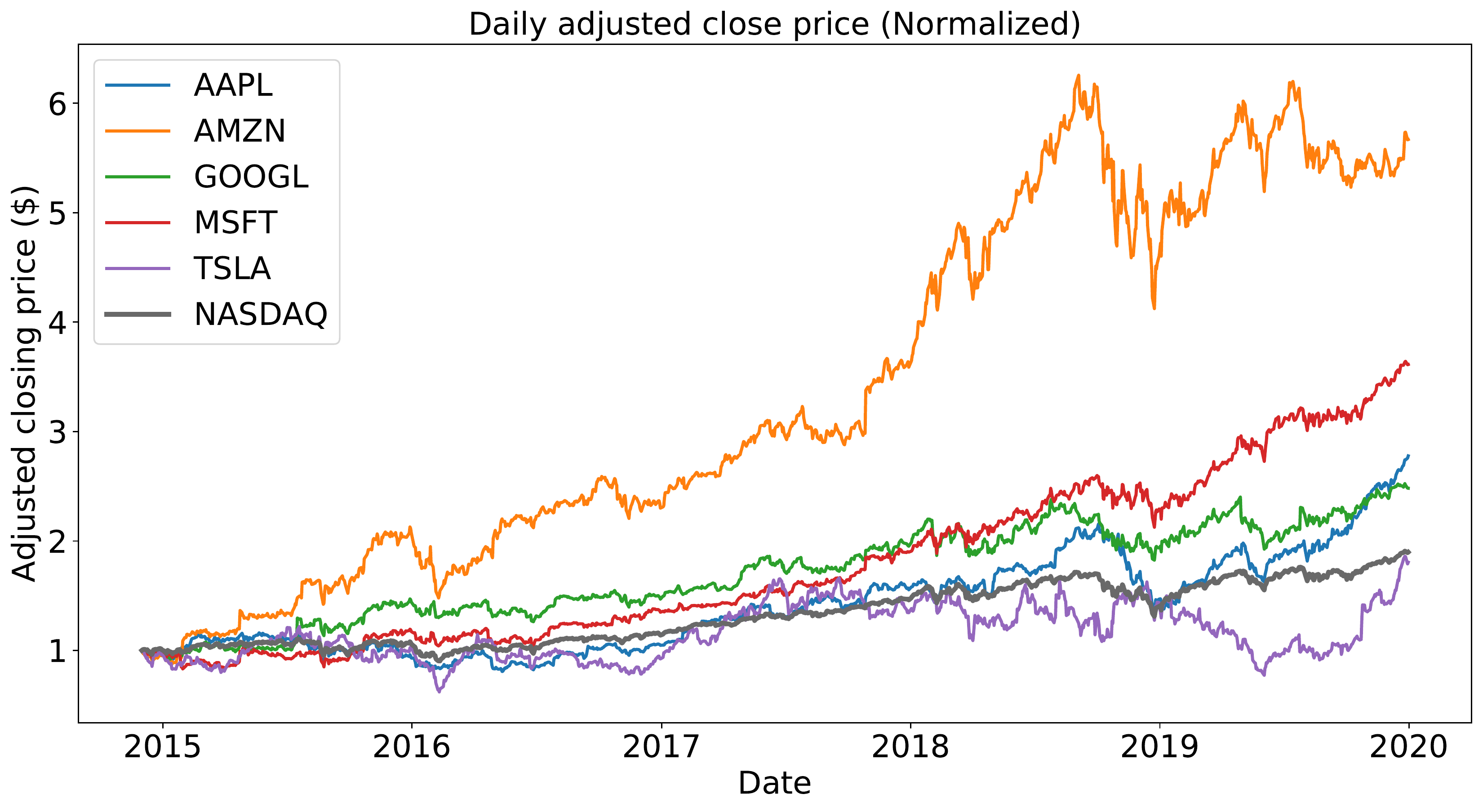}
  \centering
  \caption{Adjusted close price movements for five companies and price values of NASDAQ index for the five-year period. All prices are scaled for comparative purposes.}
  \label{fig:close_price_normalized}
\end{figure}

The financial data were collected with yfinance python library \cite{yfinance} using Yahoo! Finance's API for the same time period as the tweet data. Adjusted Close, High, Low, Open, Close, and Volume features were collected. In Figure \ref{fig:close_price_normalized}, we demonstrate the normalized adjusted close prices in order to compare the pace of each company's changes. As we can see, Amazon had the most marked growth in capitalization during the analyzed period. After analysis of Twitter data, we found out that Amazon had lower tweet activity compared to Tesla and Apple, which received much more attention from Twitter users. Despite this fact, Amazon rose nearly six times in price. Hence, we can conclude that the number of tweets is not the key feature affecting stock price movements. We need to dive into tweet semantics to get more relevant information for price movement predictions.

\subsubsection{Sentiment information}

The initial dataset with Twitter posts related to the particular group of companies did not contain information about the sentiment of the tweets. First, we had to extract sentiment polarity from the tweets in order to use them as a feature in a predictive model for comparison purposes. The following pre-trained models were applied to extract sentiment data from the tweets:

\begin{itemize}
    \item FinBERT sentiment \cite{DBLP:journals/corr/abs-1908-10063} built by further training the BERT language model on a large financial corpus and thereby adopting it for financial sentiment classification. The sentiment model is fine-tuned on 10000 manually annotated sentences collected from analyst reports about S\&P500 firms; 
    \item RoBERTa-base model fine-tuned on the Stocktwits dataset \cite{Roberta-fine-tuned}, which contains 3.2 million comments with the user labelled tags: 'Bullish' or 'Bearish'.
\end{itemize}

The latter performed better upon the comparative analysis of results for 300 tweets and was chosen for our sentiment-based solution. 


As a result of the sentiment extraction procedure, we get either 1 or 0 as labels for each tweet. Label 1 is associated with positive sentiment, while label 0 - with negative sentiment. However, before feeding obtained information into the predictive model, we need to group sentiment labels by their relation to company and date. Then, we suggest two types of scores, which reflect the ultimate sentiment of the group of tweets:
\begin{itemize}
    \item $Sentiment\;Score\;1 = negative /(negative + positive),$
    \item $Sentiment\;Score\;2 = negative / positive,$
\end{itemize}
where $Sentiment\;Score\;1$ shows the share of negative tweets in the total number of tweets, while $Sentiment\;Score\;2$ evaluates the ratio of the number of negative tweets to the number of positive tweets. $Sentiment\;Score\;2$ provided us with better metrics on the validation dataset in the task of price prediction. Thus, it is selected as the primary score for the final pipeline and is denoted $Sentiment\;Score$ in the subsequent discussion. Although, we included the behavior comparison of $Sentiment\;Score\;1$ and $Sentiment\;Score\;2$ in Section \ref{sent-explore}.

To get more a pronounced sentiment score trend instead of highly fluctuating values, three-week smoothing was applied. We compared the results of using simple rolling mean and Exponential Weighted Moving Average (EWMA) implemented in pandas library \cite{reback2020pandas}. We selected the EWMA method because of the much faster response to the downtrend in comparison with the rolling mean reaction. It happens due to the fact that EWMA algorithm contains a weight decay parameter that allows focusing more on the recent prices than on a long series of data points.  



Furthermore, we have to deal with the fact that the stock market works only during business days. As a result, there are no market data for the weekends. On the other side, sentiments have daily frequency. Options on how to unify the multiple series are the following:

\begin{enumerate}
    \item Impute market data using forward fill with previous values for gaps in a daily frequency.
    \item Convert daily frequency of tweet data to business day frequency;
\end{enumerate}

Second option was chosen in order not to introduce a series of useless values that bring no information. Sentiment score values achieved during weekends were summed up to the next business day. 

\subsubsection{Semantic information}\label{embeddings}
Another option for the representation of information from tweets is embeddings. Embedding is obtained by converting each tweet into a vector using sentence transformer model “all-MiniLM-L6-v2” \cite{all-MiniLM-L6-v2}. It maps our sentences to 384-dimensional dense vector space. To make up the representation for a single trading day, we average all vectors related to that day. In the embedding approach, we face the same problem of the absence of market prices on the weekends. To overcome that, we concatenate weekend embeddings vectors with vectors related to the next business day. 

\subsubsection{Financial information}
The predictions for future stock prices are made on the basis of historical price values. We consider open, close, high, and low prices of stocks for a particular day. Open price is a selling price of a stock at the time the exchange opens. Close price is the last price during a trading session. High and low prices are the maximum and minimum prices during a session, correspondingly. In addition, we leverage trading volume as a feature, which shows a number of shares that have been bought or sold during the trading day. 

\subsubsection{Volatility}

In the following discussion, we mean Average True Range measure under the volatility concept. Average True Range is defined according to the formula:

$$
        {ATR}_t = \frac{n-1}{n}ATR_{t-1} + \frac{1}{n}\Bigl( max(H_t, C_{t-1}) - min(L_t, C_{t-1}) \Bigr),
$$

where $t$ - given day; $ATR_t$ - Average True Range in day $t$; $n$ - considered number of periods; $H_t$ - highest price in day $t$; $L_t$ - lowest price in day $t$; $C_{t-1}$ - close price in day $t - 1$.

\subsection{Predictive model parameters}\label{model_params}

At the stage of selecting architecture for price prediction purposes, multiple models implemented in Darts \cite{JMLR:v23:21-1177} python library were tested. The best performance showed the implementation of Temporal Fusion Transformer \cite{https://doi.org/10.48550/arxiv.1912.09363}, which was chosen to become the core predictive model in the final pipeline.  
Instead of a univariate approach that implies the creation of a distinct model for each company, multiple time-series training is considered. This allows to train multiple time-series objects in one go, simplifying the training and forecasting process. 

As the input history length of the model, previous three business weeks (15 days) were chosen. Such length for the lookback window was achieved during grid-search and mentioned in the Appendix. The shorter lookback windows introduce certain limitations. In particular, the sentiment reaction to the event requires time, and extreme shortening of the observed history may lead to information loss. The output prediction length was set equal to 3 days for the main experiments. For the train/test split 80\% was given for the training dataset and 20\% for testing.

A common target variable in all experiments is close price. To test an effect of different input compositions, three groups were identified:

\begin{itemize}
    \item HLOV - High, Low, Open, Volume attributes. In this case, we test out the model performance only working with market data as input.
    \item HLOVS - the same as above, but with adding $Sentiment\;Score$ attribute.
    \item HLOVE - market data and embedding vectors as model input
\end{itemize}

\subsection{Loss function}
In order to prevent overfitting, a custom loss function is introduced. It is built on top of one of the most popular loss functions for regression tasks - Mean Squared Error (MSE) loss, but with the addition of a directional component. MSE does not consider whether the direction of the predicted close price is correct, just focusing on the difference between true and predicted prices. However, the direction of price movement is an essential factor in the financial world. Multiple values from 10 to $10^4$ of $\alpha$ were tested in the loss function. Lower values of $\alpha$ did not punish the directional errors enough to make a difference in comparison with standard MSE loss, while having $\alpha$ set to $10^4$ only prolonged the loss convergence time. In the end, we come up with the following loss function:
$$
DMSE = \frac{1}{n}\sum_{i=1}^{n}\alpha(x_i-y_i)^2, \quad \text{where} \hspace{0.5em}
\begin{cases}
    \alpha = 1,& \text{if } (x_i - x_{i-1})(y_i - y_{i-1}) \geq 0,\\
    \alpha = 10^3,              & \text{otherwise},
\end{cases}
$$


where $DMSE$ - Directional Mean Squared Error; $n$ - number of data points; $x_{i}$ - true values; $y_{i}$ - predicted values.

%% file: sections/results.tex
\section{Results}

In this section, we present the findings from our study on the relationships between Twitter sentiments, stock prices, and volatility. Also, we perform metrics achieved from predictive model development using sentiment labels and sentence embeddings. 

\subsection{Daily return analysis}

\begin{figure}[!ht]
  \includegraphics[width=0.9\textwidth]{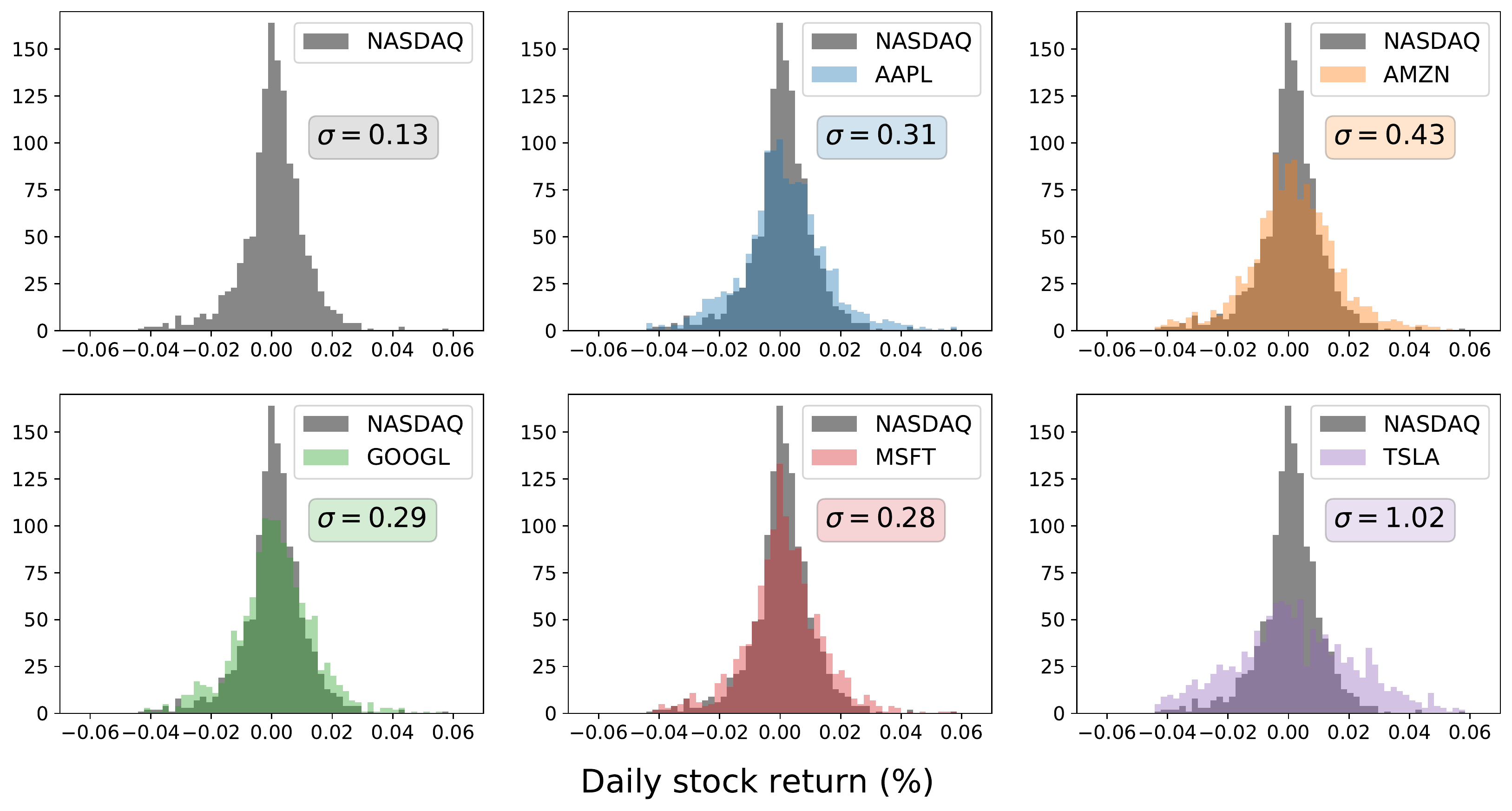}
  \centering
  \caption{Histograms of daily stock return values for the companies during a five-year period. NASDAQ daily stock returns are given for comparison purposes. Tesla is the most volatile among the analyzed five companies.}
  \label{fig:return distribution}
\end{figure}

Histograms of daily stock return values for all five companies in a five-year period are presented in Figure \ref{fig:return distribution}. Here $\sigma$ is calculated as the sum of squares of deviations of daily return values from 0. As we can observe, Tesla is a clear outlier among five other companies having much greater daily return oscillations. One can infer that this stock has a higher risk/reward ratio. This shows that no matter that all the described stocks are traded inside the same NASDAQ index, there is still evidence of them behaving in a different way caused by the other factors.

\subsection{Relation between tweet sentiments and stock market}\label{sent-explore}

\begin{figure}[!ht]
    \centering
    \subfigure[Apple]{\includegraphics[width=8cm]{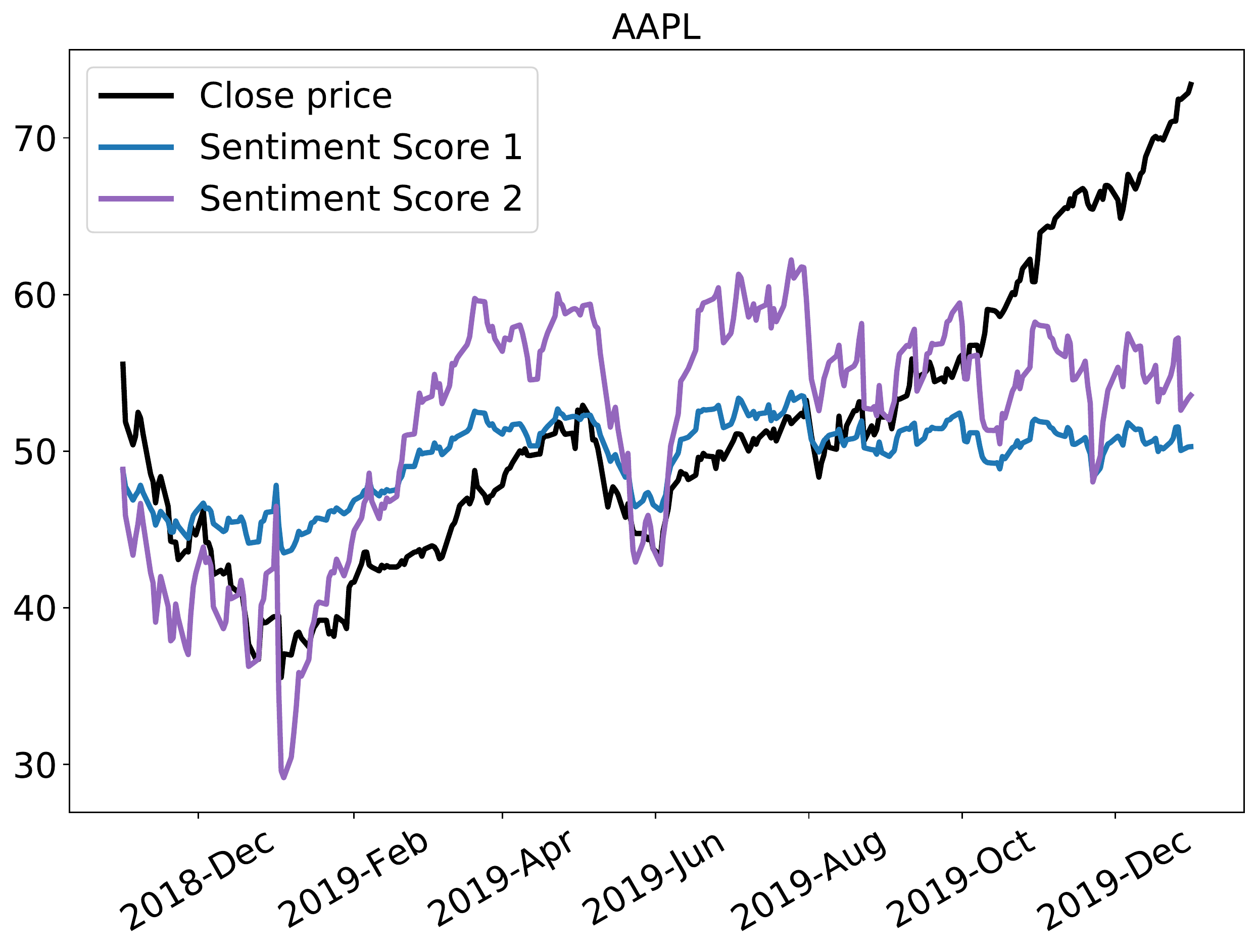}}
    \subfigure[Amazon]{\includegraphics[width=8cm]{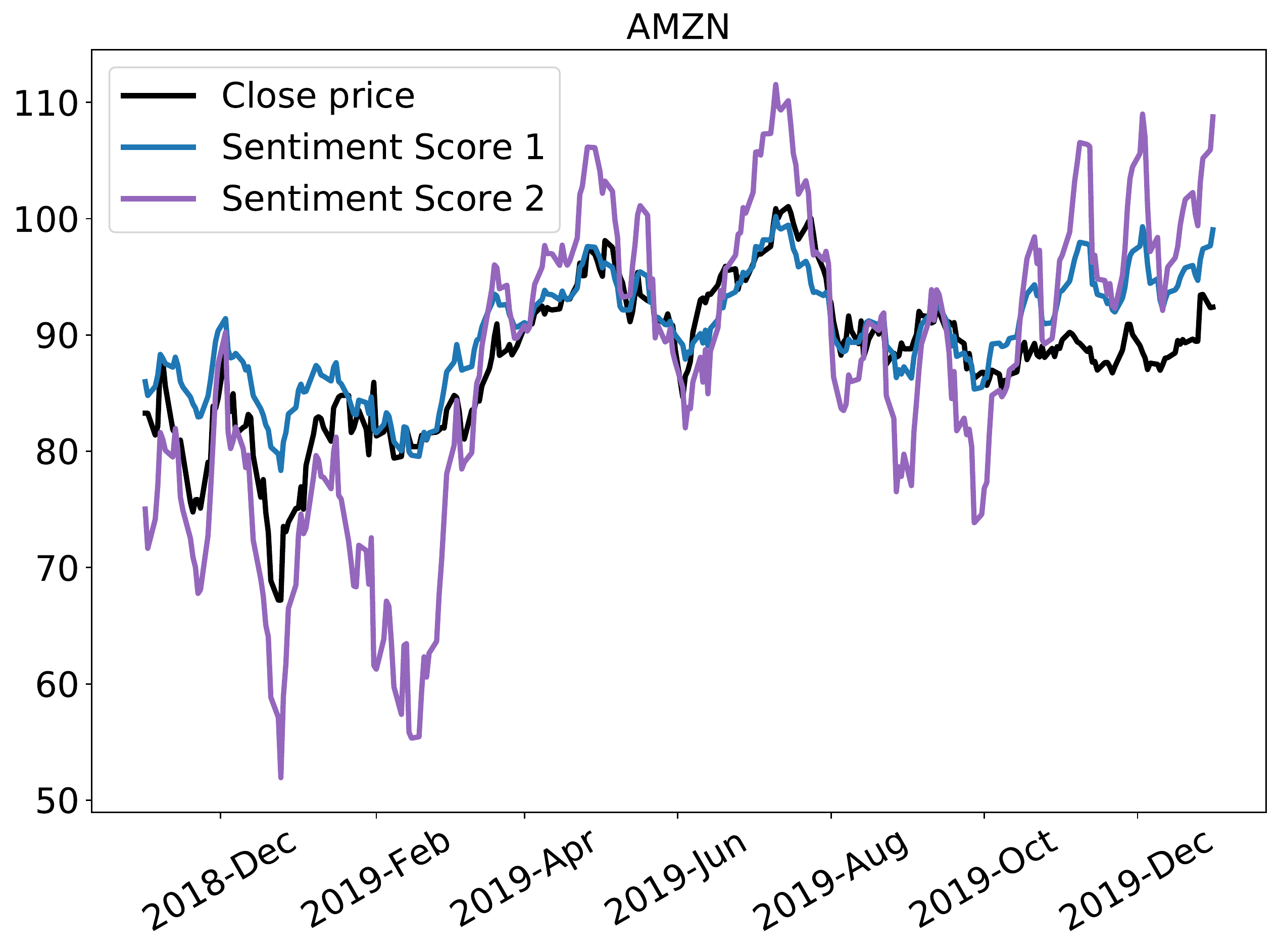}}
    \caption{Sentiment scores and closing prices comparison with a constant multiplier applied to sentiment scores for visualization purposes. Left side is denoted to the moving of Apple prices and sentiments, while Amazon patterns are demonstrated on the right side.}
    \label{fig:sent_price}
\end{figure}

Before the construction of the predictive model on the basis of Twitter data, our goal is to explore the direct connection between price variables and Twitter sentiments. Figure \ref{fig:sent_price} demonstrates stock prices as well as sentiment scores calculated in both ways for Apple and Amazon. Visually, we can observe a great amount of resemblance between price and sentiment variables, especially in the case of Amazon. 

\begin{figure}[!ht]
    \centering
    \subfigure[Sentiment score and volatility behavior]{\includegraphics[width=8cm]{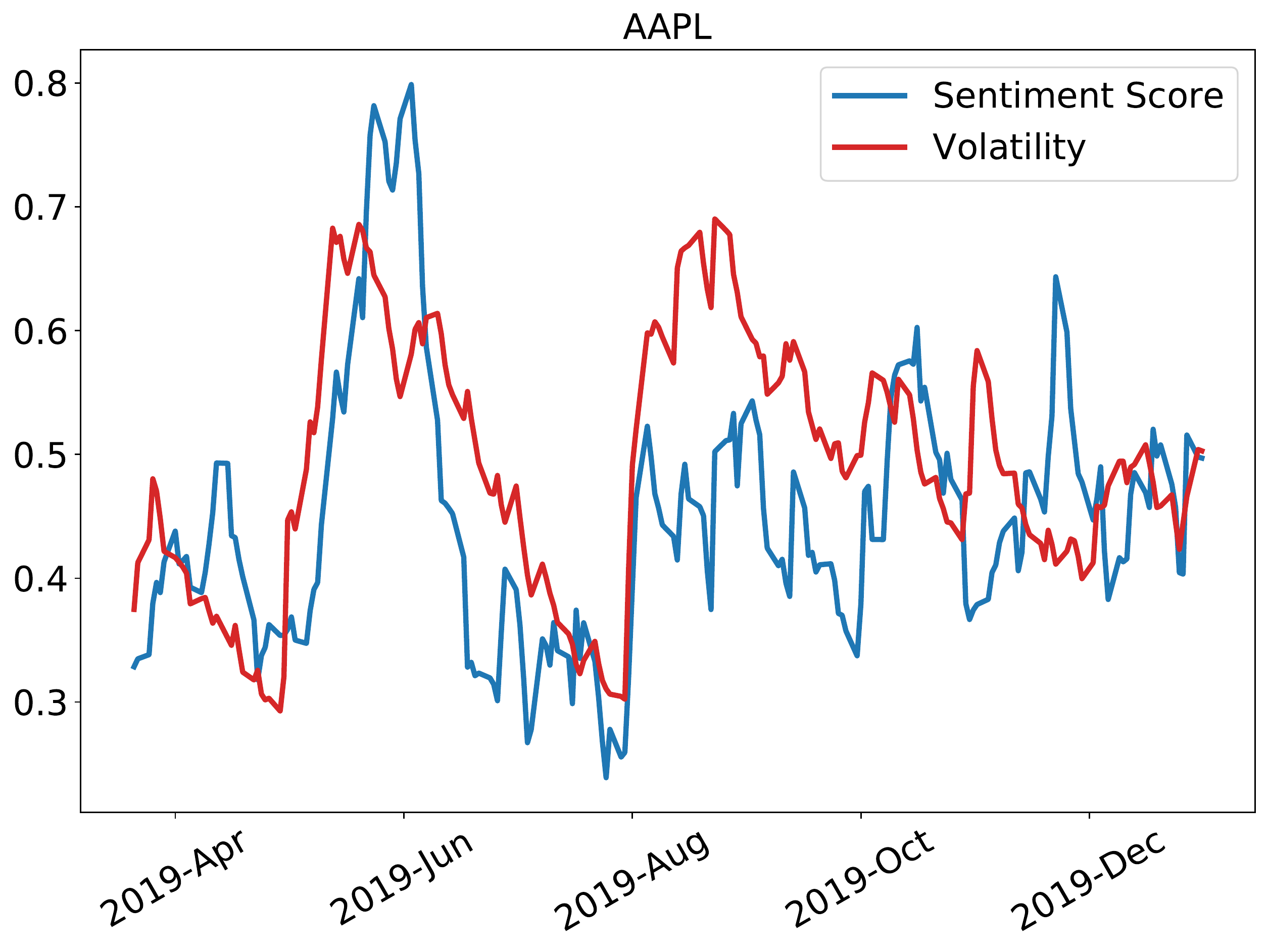}}
    \subfigure[Scatter plot for sentiment score and volatility]{\raisebox{4.2mm}{\includegraphics[width=8cm]{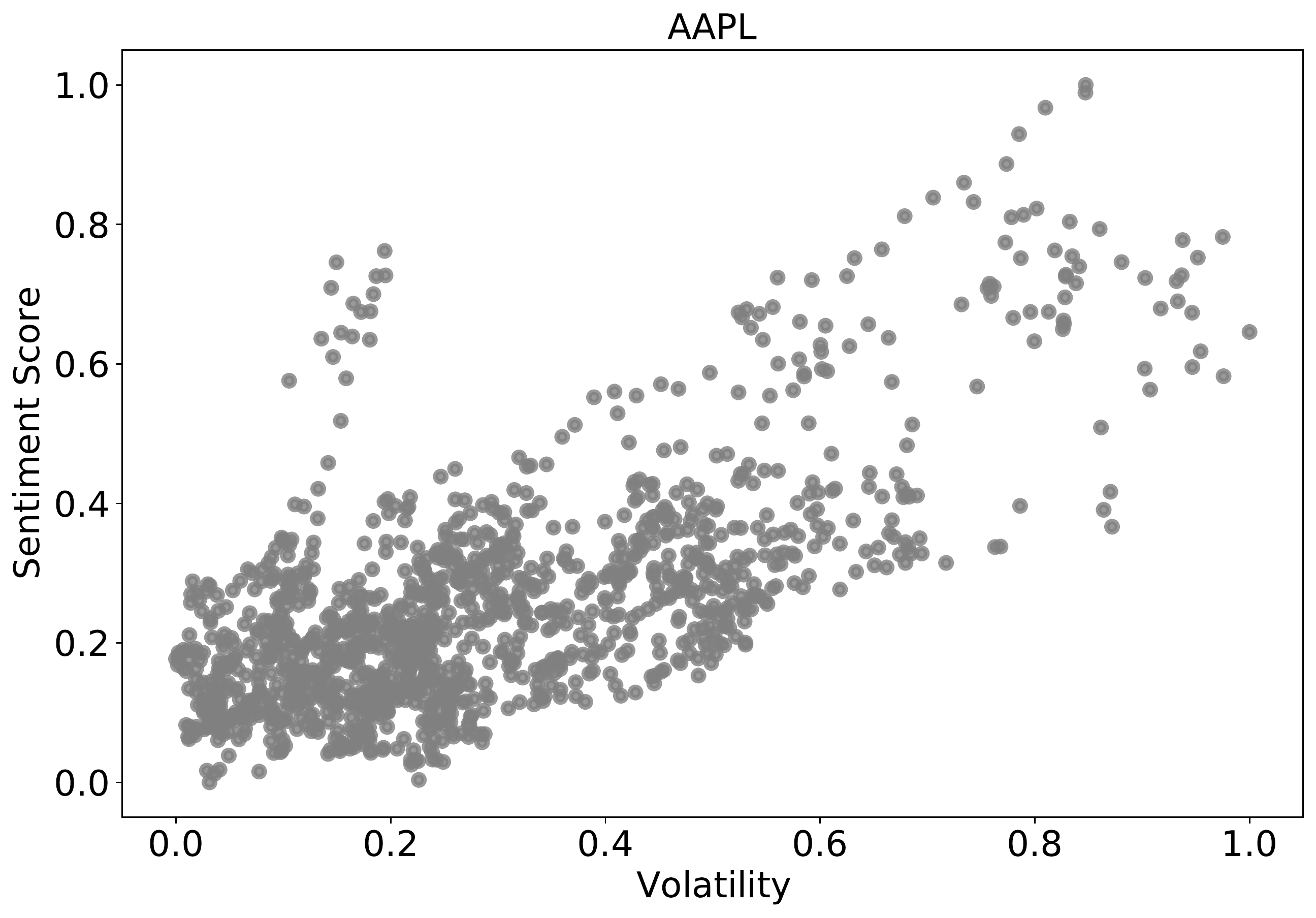}}}
    \caption{Sentiment score and volatility comparison for Apple with Min-Max scaling applied to both features. There is a comparison of behaviour of sentiment scores and volatility overtime on the left side. We can observe a time lag of sentiment reaction to volatility. The right side is denoted to the scatter plot of sentiment score and volatility to discover hidden patterns.}
    \label{fig:sent_volatility}
\end{figure}

\begin{figure}[!ht]
    \centering
    \subfigure[Sentiment score and volatility behavior ]{\includegraphics[width=8cm]{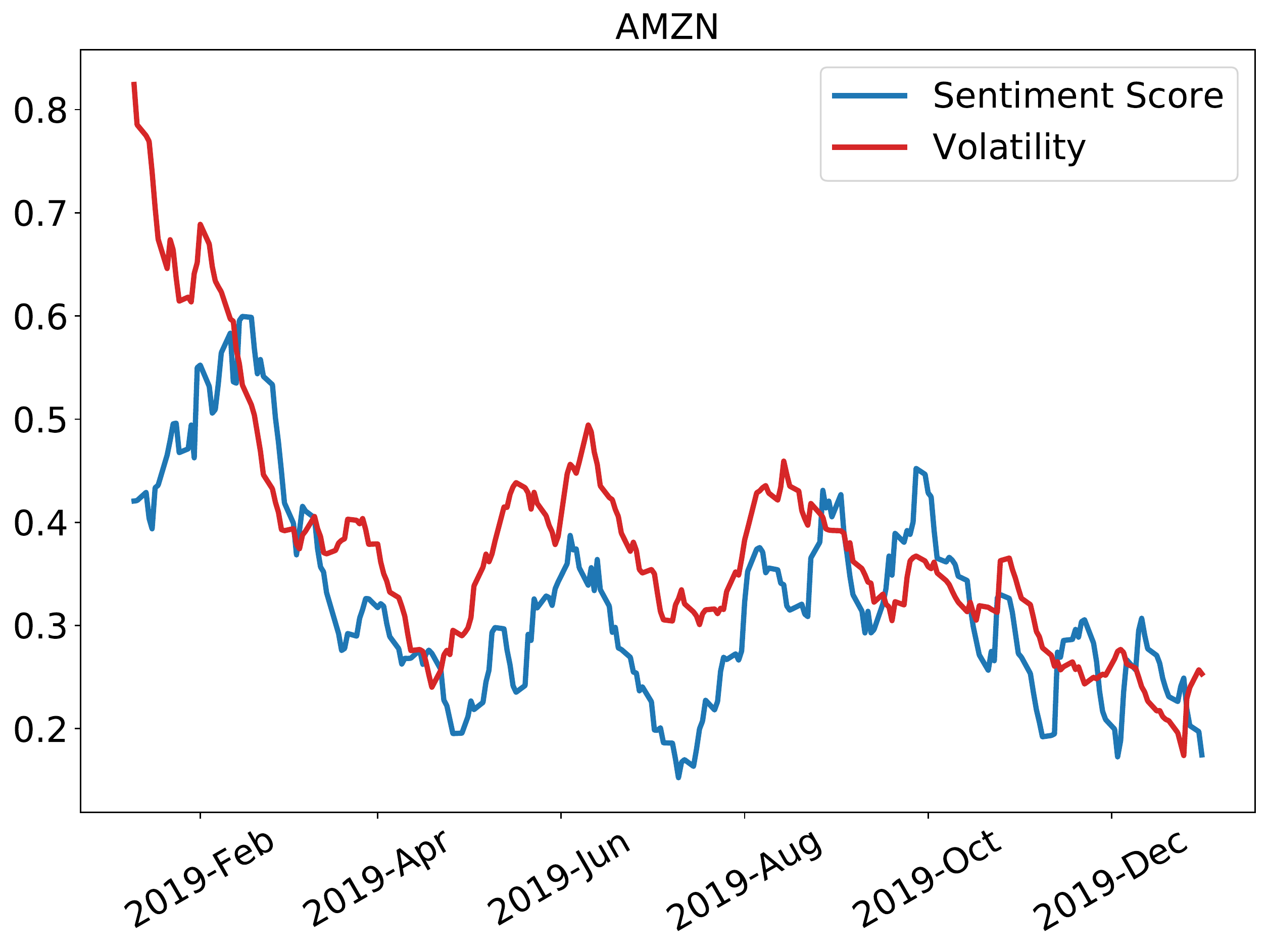}}
    \subfigure[Scatter plot for sentiment score and volatility]{\raisebox{4.2mm}{\includegraphics[width=8cm]{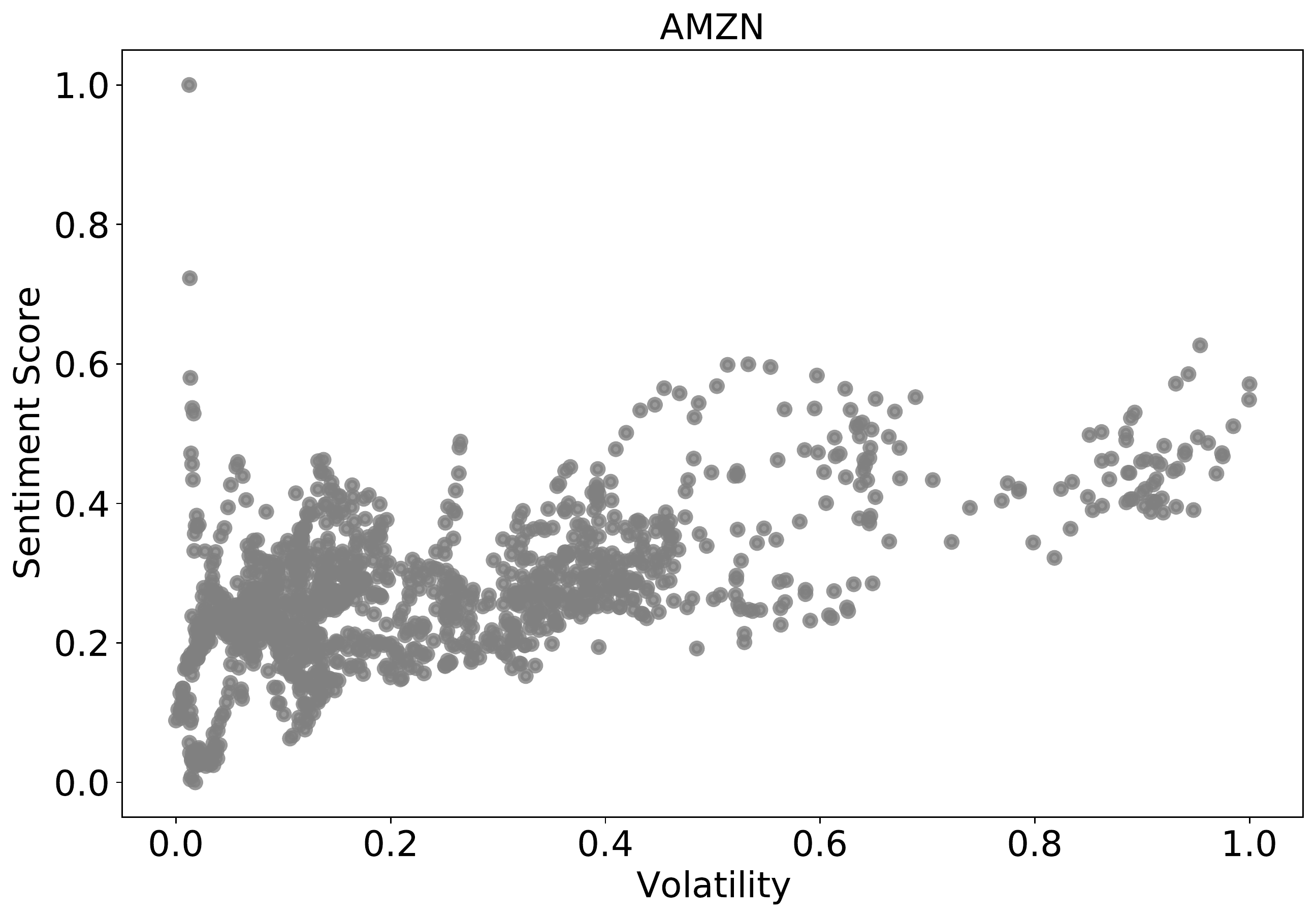}}}
    \caption{Sentiment score and volatility comparison for Amazon with Min-Max scaling applied to both features. There is a comparison of behaviour of sentiment scores and volatility overtime on the left side. We can observe quite synchronized directional movements. The right side is denoted to the scatter plot of sentiment score and volatility to explore hidden patterns in more detail.}
    \label{fig:sent_volatility_scatter}
\end{figure}

If we compare stock volatility with EWMA smoothed sentiment score in Figures \ref{fig:sent_volatility} and \ref{fig:sent_volatility_scatter}, we observe the same phenomenon of significant similarity but now between the shape of volatility curve and Twitter sentiment scores. Amazon again has a greater correlation between considered values. It is noteworthy that Apple has a clear time lag between public reaction and stock volatility. Volatility tends to precede public sentiment response. For Amazon, in turn, the situation is slightly different. Sentiment changes are more synchronized with the volatility making it a greater predictor of price movement. Analyzing scatter plots in Figures \ref{fig:sent_volatility} and \ref{fig:sent_volatility_scatter}, we can ascertain the presence of principle direction in sentiment-volatility dependence.

To conduct a more comprehensive correlation analysis between different variables, we consider Google company and calculate pairwise Spearman correlation coefficients between close price, volume, volatility, and sentiment score. The results are given in Table \ref{table:EMA_correlation}. We applied three-week EWMA smoothing to volume and sentiment score in order to get a more meaningful correlation with the stock close price. After comparing with non-smoothed calculations, we found out that smoothing indeed helped to improve correlation indicators. It happens because we smoothed out short-term fluctuations and consequently captured the underlying trends.

\begin{table}[!ht]
    \centering
    \caption{Spearman correlation between different variables with and without application of three-week window smoothing to volatility and sentiment score}
    \begin{tabular}{ccccc}
    \toprule
    Using EWMA & Close & Volume & Volatility & Sentiment Score \\
    \midrule
    Close &  1.000 & -0.464 &  0.626 & 0.401 \\
    Volume & -0.464 &  1.000 & 0.295 & 0.085 \\
    Volatility  &  0.626 &  0.295 & 1.000 & 0.508 \\
    Sentiment Score & 0.401& 0.084 & 0.508 & 1.000 \\
    \bottomrule
    \end{tabular}
    \label{table:EMA_correlation}
\end{table}


\subsection{Connection of sentiments and embeddings}
After exploring of relationships between sentiment scores and financial indicators, we became interested in the interdependency of other entities, namely tweets' sentiments and their embeddings. 
As mentioned, tweet sentiments were extracted with a pre-trained RoBERTa-base model, and tweet embeddings were generated by sentence transformers. We hypothesize the presence of an underlying connection between these two representations. To check this, we fitted a linear regression model on vector embeddings, for which corresponding sentiment scores were set as targets. The prediction quality was estimated with $R^2$ metric. The results of such experiment are given in Table \ref{tab:sklearn-table}. For a better understanding of the magnitude of the effect that embedding encompasses sentiment knowledge, we generate vectors of random variables with a dimension equal to the 
size of embedding vectors. Then, we try to predict sentiment scores from obtained pointless vectors. 




\begin{table}[!ht]
    \centering
    \caption{$R^2$ values for each company achieved after fitting a linear regression model on embeddings as features and sentiment scores as targets. $R^2$ scores for the case when predictions are made from randomly generatedvectors are given for comparison purpose.}
    \begin{tabular}{cccccc}
        \toprule
       $R^2$ for $Sentiment\:Score$ prediction &  Apple &  Amazon &  Google & Microsoft & Tesla \\
          from embeddings (random vectors) &  0.800 (0.344) & 0.821 (0.352) & 0.778 (0.358) & 0.804 (0.381) & 0.891 (0.396) \\
   
        \bottomrule
    \end{tabular}
    \label{tab:sklearn-table}
\end{table}

For the majority of companies, $R^2$ values are greater than 0.8. Thus, there is a strong relationship between the vector embeddings and the sentiment scores. High metrics for the linear regression model indicate that the feature variable is a strong predictor of the response variable. 


\subsection{Price prediction results and sensitivity analysis}\label{prediction_results}

When choosing price prediction models to include in the final pipeline, several candidates were tested. Moreover, we experimented with different horizons. All details and results of intermediate experiments are given in Appendix. For the main experiments, we took the model and forecast horizon that proved themselves in the best way during the selection phase. We stopped at the TFT model with the forecast horizon equal to 3.


The aim of the experiments is to compare metrics for a three-day prediction period when using sentence embedding or sentiments as features. We basically compare the effect by considering two feature sets: HLOVS and HLOVE, which are described in \ref{model_params}. The metrics are performed in Table \ref{table:grouped-error-table}. As we wanted to make a generalizable conclusion, we examined embeddings that are generated by one more language model. We considered Microsoft model - all-mpnet-base-v2 \cite{all-mpnet-base-v2}. It has twice the number of dimensions compared with the approach mentioned in \ref{embeddings}. The feature set with embeddings obtained with this model is denoted as HLOVE2 in Table \ref{table:grouped-error-table}. However, the accuracy of closing price prediction dropped significantly for such embeddings.

\begin{table}[!ht]
    \centering
    \caption{Metrics for three-day forecast horizon considering five companies. The predictions are made with TFT model using feature sets including either sentiments (HLOVS) along with financial data or embeddings (HLOVE, HLOVE2).}
    \begin{tabular}{llccccccc}
    \toprule
          &  Model          &    MAPE &     MAE &      $R^{2}$ &    RMSE &     MSE &   SMAPE \\
    \midrule
        Apple & \textbf{tft\_HLOVS} &        1.7602 &  \textbf{0.8932} &  \textbf{0.9648} &  \textbf{1.2014} &  \textbf{1.4434} &  1.7502 \\
              & tft\_HLOVE &        \textbf{1.7557} &  0.8947 &  \textbf{0.9648} &  1.2023 &  1.4455 &  \textbf{1.7492} \\
              & tft\_HLOVE2 &       3.9357 &  1.6854 &  0.7368 &  2.1665 &  4.6938 &  3.8641 \\
              \\
        Amazon & \textbf{tft\_HLOVS} &       \textbf{1.0544} &  \textbf{0.5605} &  \textbf{0.8535} &  \textbf{0.7469} &  \textbf{0.5579} &  \textbf{1.0526} \\
              & tft\_HLOVE &       1.1442 &  0.6078 &  0.8334 &  0.7965 &  0.6344 &  1.1408 \\
              & tft\_HLOVE2 &      1.9503 & 1.0404 & 0.4551 & 1.5635 &  2.4446 & 1.9184 \\
              \\
        Google & \textbf{tft\_HLOVS} &       1.6830 &  0.9044 &  \textbf{0.8326} &  \textbf{1.2160} &  \textbf{1.4786} &  1.6807 \\
              & \textbf{tft\_HLOVE} &      \textbf{1.6824} &  \textbf{0.9039} &  0.8323 &  1.2171 &  1.4814 &  \textbf{1.6781} \\
              & tft\_HLOVE2 &       2.9347 & 2.3175 & 0.4988 & 2.6873 & 2.9715 & 3.3174 \\
              \\
        Microsoft & \textbf{tft\_HLOVS} &       \textbf{1.2687} &  \textbf{0.8267} &  \textbf{0.9626} &  \textbf{1.0643} &  \textbf{1.1327} &  \textbf{1.2721} \\
              & tft\_HLOVE &      1.3059 &  0.8503 &  0.9615 &  1.0798 &  1.1660 &  1.3086 \\
              & tft\_HLOVE2 &       3.9615 & 1.6081 & 0.7412 & 3.5431 & 2.9846 & 3.0965 \\
              \\
        Tesla & \textbf{tft\_HLOVS} &       \textbf{3.5182} &  \textbf{1.3833} &  \textbf{0.8923} &  \textbf{1.9810} &  \textbf{3.9242} &  \textbf{3.5386} \\
              & tft\_HLOVE &       3.5717 &  1.4020 &  0.8915 &  1.9880 &  3.9523 &  3.5994 \\
              & tft\_HLOVE2 &       4.5540 & 1.9531 & 0.5735 & 2.3948 &  5.7354 &  4.4676 \\
    \bottomrule
    \end{tabular}
    \label{table:grouped-error-table}
\end{table}

If we look at prediction qualities for Apple company in Table \ref{table:grouped-error-table}, we observe that for some metrics embedding approach shows better performance. However, the difference from the sentiment approach is not substantial, and HLOVS set gives better results for 4 out of 6 metrics. For Microsoft, sentiment representations clearly perform better. Another company that has nearly similar results for both feature sets is Google, and the difference in metrics is negligible - only 0.036\% for MAPE. For Amazon and Tesla sentiments again prove to be an optimal feature for all metrics. To sum up, although embeddings seem to contain more information, sentiment solution remains a strong baseline that is hard to beat. 

If we analyze MAPE metric, models with embedding feature show better results only in two cases out of five - for Apple and Google. However, this prevalence is minor. During the analysis of volatility and close price correlations with sentiment score, these companies had one of the lowest scores among the considered five. It might be the reason why Twitter sentiments do not have such a performance-enhancing impact for these two companies. For Amazon, Google and Tesla sentiment score method outperforms embedding vector solutions.

%% file: sections/conclusions_revised.tex
\section{Conclusions}

This study provides evidence that information obtained from Twitter posts serves as a strong indicator for stock price movements. In particular, Twitter sentiment scores are highly correlated with price volatility and improve the performance of predictive models for financial markets. We conclude that sentiments provide valuable insights on events around and contribute to better capturing underlying market dynamics. 

The main goal of this paper is to investigate whether sentence embeddings yield better results in stock price prediction compared to the sentiment analysis approach. In the majority of conducted experiments, the sentiment approach outperforms the embedding vectors method. This fact might be counterintuitive because embeddings seem to encompass more valuable contextual information. However, sentiments tend to represent information in a more concise way, bringing less noise into the prediction model. Nevertheless, the embedding approach still has an advantage that it does not require an additional model for sentiment extraction and the consequent quality verification of that procedure. It is important to note the limitation of the conducted research. We made a comparison analysis only within restricted use case of Twitter posts and stocks of top companies from NASDAQ.  

\section{Statement on computational resources and environmental impact}  

We used a NVIDIA GeForce RTX 3080 Ti GPU to train the models, extract the sentiment score and make embedding vectors from the tweets using BERT model \cite{all-MiniLM-L6-v2}. Two NVIDIA A100 80GB PCIe GPUs were used for testing out the MPNet \cite{all-mpnet-base-v2} approach and inferencing. This work contributed totally 6.12 kg equivalent $CO_{2}$ emissions. The carbon emissions information was generated using the open-source library \textit{eco2AI}\footnote{Source code for \textit{eco2AI} is available at \url{https://github.com/sb-ai-lab/Eco2AI}} \cite{budennyy2023eco2ai}. 

%% file: sections/annex.tex
\section*{Appendix}

\subsection*{Hyperparameters}

Listed hyperparameters were used in gridsearch procedure. Finally chosen parameters are highlighted in bold.

Hyperparameters used for TFT training:
\begin{itemize}
    \item lookback window = 5, \textbf{15}, 32
    \item hidden\_size = 15, 32, \textbf{64}, 80 - the main hyper parameter among TFT architecture, which describes the number of neurons of each dense layer during variable selection process, static enrichment section and position-wise feed forward;
    \item lstm\_layers = \textbf{1}, 2 - number of layers for the LSTM encoder and decoder;
    \item num\_attention\_heads = 2, \textbf{4} -  number of attention heads;
    \item feed\_forward = ReLU, \textbf{SwiGLU} - according to the paper \cite{https://doi.org/10.48550/arxiv.2002.05202}, GLU activations improve transformer-based architecture performance that proved to be the case in our problem as well;
    \item dropout = 0.10, 0.15, \textbf{0.25}, 0.5 - fraction of neurons that are affected by dropout;
    \item hidden\_continuous\_size = 15, \textbf{32}, 64 - hidden size for processing continuous variables;
    \item norm\_type = LinearNorm, \textbf{RMSNorm} - a simplification of the original layer normalization;
    \item optimizer\_cls = \textbf{Adam}, AdamW, Adagrad - standard optimization algorithm;
    \item batch\_size = 8, 16, \textbf{32}, 64.
\end{itemize}

Hyperparameters used for Nlinear training:
\begin{itemize}
    \item lookback window = 5, \textbf{15}, 32
    \item const\_init = False, \textbf{True} - initialize the weights to $1 / input\_length$ instead of default PyTorch initialization;
    \item optimizer\_cls = \textbf{Adam}, AdamW, Adagrad - standard optimization algorithm;
    \item batch\_size = 8, 16, \textbf{32}, 64.
\end{itemize}

\subsection*{Data preprocessing}

We have a total of 3717964 tweets in the dataset from 140131 writers. This means that a significant amount of tweets were written by the same accounts through the period from the beginning of 2015 to the end of 2019, and that is why during the data preparation step, we had to deal with duplicate tweets. Less than 1\% of NaN values were observed only in the writer column (deleted accounts) and were dropped.

First, we removed the tweets with more than one stock mentioned in the body of a tweet to avoid ambiguous cases, which resulted in the removal of 421101 tweets – 11.3\% of the total amount. Then we drop intra-day duplicate tweets generated by the bot accounts by complete matches in the body, getting rid of another 4.9\% of remaining tweets. After that, a text cleaning function was applied, removing web links, mentions, tickers, unnecessary punctuation, revealing another 11.5\% of duplicate tweets, getting to about 2.7 million total tweets.

\subsection{Selection of prediction model and sensitivity analysis}

We considered several architectures when choosing a close price prediction model. As the baseline solution, we took the Naive Seasonal model, which returned the close price for $N$ steps ahead. Also, we experimented with N-Linear predictive model. 

The results for each company, feature set, and model are demonstrated in Table \ref{table:total-error-table}. Three-day and five-day forecast horizons were tested. The three-day window performs better, so the corresponding results are included in the main part of this paper. The five-day model output gives worse metrics but is worth analysis. Metrics in Table \ref{table:total-error-table} are calculated for five days ahead prediction. There is a tendency for TFT to be more capable of extracting the information from the embedding vector compared to N-Linear.

For Apple, the best performing model is TFT with embedding vector as a feature. The usage of sentiment score does not help to improve the model's accuracy for both TFT and N-Linear. Only two model instances outperform the baseline Naive Seasonal approach - N-Linear with sentiment score and embedding vector.

For Amazon, we observe different behaviour of the model's performance given the input features. Both N-Linear and TFT receive performance boost with the utilization of sentiment score. Embedding vector yields higher error values. Even the baseline solution outperforms the embedding approach by a significant margin. Thus, no useful data are extracted from this feature in this case.

Tesla is the most volatile stock amount the five considered. It is evident that all models struggle with stock price prediction for Tesla, bringing the highest error values among all stocks. Only TFT in combination with the sentiment score performs better than the naive baseline solution. All of the other models and features show worse accuracy.

\begin{table}[!ht]
    \centering
    \caption{Metrics for five days ahead prediction experiments. Different input feature sets are considered. For each company, the composite ranking takes into account multiple measures of model performance.}
    \begin{tabular}{llcccccc}
    \toprule
          Company & Model &  Rank & MAPE & MAE &      $R^{2}$ &    RMSE &   SMAPE \\
    \midrule
          Apple & baseline  &     5 &   2.5654 &  1.3132 &  0.9344 &  1.6736 &  2.5867 \\
          & nlinear\_HLOV &     4 &   2.5506 &  1.3033 &  0.9280 &  1.7074 &  2.5489 \\
          & nlinear\_HLOVS &     6 &  2.6844 &  1.3622 &  0.9253 &  1.7383 &  2.6814 \\
          & nlinear\_HLOVE &     7 &   4.4765 &  2.2690 &  0.8050 &  2.8094 &  4.4628 \\
          & tft\_HLOV &     2 &   2.4696 &  1.2583 &  0.9383 &  1.5796 &  2.4805 \\
          & tft\_HLOVS &     3   &   2.4981 &  1.2772 &  0.9292 &  1.6928 &  2.4892 \\
          & \textbf{tft\_HLOVE} &     1 &   \textbf{2.1485} &  \textbf{1.0922} &  \textbf{0.9502} &  \textbf{1.4203} &  \textbf{2.1468} \\ \hline
          \\
          Amazon & baseline  &     3 &   1.4832 &  0.7907 &  0.7235 &  1.0273 &  1.4837 \\
          & nlinear\_HLOV &     6 &   1.6984 &  0.9044 &  0.6192 &  1.2135 &  1.7003 \\
          & nlinear\_HLOVS &     5 &  1.6619 &  0.8865 &  0.6403 &  1.1795 &  1.6650 \\
          & nlinear\_HLOVE &     7 &   3.7163 &  1.9763 & -0.6065 &  2.4926 &  3.7144 \\
          & tft\_HLOV &     2 &   1.3785 &  0.7356 &  \textbf{0.7622} &  \textbf{0.9589} &  1.3780 \\
          & \textbf{tft\_HLOVS} &     1 &  \textbf{1.3161} &  \textbf{0.7005} &  \textbf{0.7622} &  0.9590 &  \textbf{1.3129} \\
          & tft\_HLOVE &     4 &   1.6166 &  0.8594 &  0.6640 &  1.1399 &  1.6013 \\  \hline
          \\
           Google & baseline  &     4 &   2.2939 &  1.2319 &  0.7119 &  1.6087 &  2.2950 \\
          & nlinear\_HLOV &     6 &   2.8414 &  1.5251 &  0.4790 &  2.1344 &  2.8412 \\
          & nlinear\_HLOVS &     5 &  2.8261 &  1.5128 &  0.4703 &  2.1523 &  2.8240 \\
          & nlinear\_HLOVE &     7 &   5.3571 &  2.9040 & -0.5161 &  3.6410 &  5.3650 \\
          & tft\_HLOV &     3 &   2.0767 &  1.1163 &  0.7479 &  1.4848 &  2.0781 \\
          & \textbf{tft\_HLOVS} &     1 &  \textbf{2.0209} &  \textbf{1.0844} &  \textbf{0.7565} &  \textbf{1.4592} &  \textbf{2.0192} \\
          & tft\_HLOVE &     2 &   2.0657 &  1.1022 &  0.7088 &  1.5956 &  2.0486 \\ \hline
          \\
          Microsoft & baseline  &     4 &   1.8159 &  1.1821 &  0.9303 &  1.4773 &  1.8287 \\
          & nlinear\_HLOV &     5 &   2.0021 &  1.3021 &  0.9119 &  1.6424 &  2.0035 \\
          & nlinear\_HLOVS &     6 &  2.0701 &  1.3441 &  0.9072 &  1.6850 &  2.0715 \\
          & nlinear\_HLOVE &     7 &   3.7334 &  2.4173 &  0.6771 &  3.1433 &  3.7348 \\
          & tft\_HLOV &     3 &   1.7392 &  1.1442 &  0.9294 &  1.4703 &  1.7546 \\
          & tft\_HLOVS &     2 &   \textbf{1.6815} &  1.1088 &  0.9311 &  1.4521 &  1.6937 \\
          & \textbf{tft\_HLOVE} &     1 &   1.6936 &  \textbf{1.0969} &  \textbf{0.9359} &  \textbf{1.4003} &  \textbf{1.6819} \\ \hline
          \\
          Tesla & baseline &     2 &   5.1027 &  2.0321 &  0.7996 &  2.7298 &  5.0911 \\
          & nlinear\_HLOV &     5 &   5.9677 &  2.4198 &  0.7107 &  3.2337 &  5.9809 \\
          & nlinear\_HLOVS &     6 &  5.9770 &  2.4363 &  0.7090 &  3.2432 &  5.9771 \\
          & nlinear\_HLOVE &     7 &   6.4840 &  2.5655 &  0.6965 &  3.3120 &  6.4805 \\
          & tft\_HLOV &     4 &   5.2499 &  2.1137 &  0.7462 &  3.0284 &  5.1821 \\
          & \textbf{tft\_HLOVS} &     1 &  \textbf{4.8706} &  \textbf{1.9418} &  \textbf{0.8143} &  \textbf{2.5904} &  \textbf{4.8720} \\
          & tft\_HLOVE &     3 &   5.1824 &  2.0705 &  0.7840 &  2.7940 &  5.1357 \\  
    \bottomrule
    \end{tabular}
    \label{table:total-error-table}
\end{table}
